\pdfoutput=1

\documentclass[11pt]{article}
\usepackage{multirow}
\usepackage{enumitem}
\usepackage{xcolor}
\usepackage{xspace}
\usepackage{comment}
\usepackage{verbatim}
\usepackage{mdframed}
\usepackage{amsmath}
\usepackage{float}

\usepackage{subcaption}
\usepackage[preprint]{acl}

\usepackage{times}
\usepackage{latexsym}
\usepackage{booktabs}
\usepackage{graphicx}
\usepackage{amsmath}
\usepackage[T1]{fontenc}

\usepackage[utf8]{inputenc}

\usepackage{microtype}

\usepackage{inconsolata}

\usepackage{graphicx}
\newcommand{\method}{\texttt{SECRET}\xspace}

\newcommand{\blockcomment}[1]{}

\title{\method: Semi-supervised Clinical Trial Document Similarity Search}

\author{
  \textbf{Trisha Das\textsuperscript{1}},
  \textbf{Afrah Shafquat \textsuperscript{2}},
  \textbf{Beigi Mandis\textsuperscript{2}},
  \textbf{Jacob Aptekar\textsuperscript{2}},
\\
  \textbf{Jimeng Sun \textsuperscript{1}},
\\
  \textsuperscript{1}University of Illinois Urbana-Champaign,
  \textsuperscript{2}Medidata Solutions,
}

\begin{document}
\maketitle
\begin{abstract}
Clinical trials are vital for evaluation of safety and efficacy of new treatments. However, clinical trials are resource-intensive, time-consuming and expensive to conduct, where errors in trial design, reduced efficacy, and safety events can result in significant delays, financial losses, and damage to reputation. These risks underline the importance of informed and strategic decisions in trial design to mitigate these risks and improve the chances of a successful trial. Identifying similar historical trials is critical as these trials can provide an important reference for potential pitfalls and challenges including serious adverse events, dosage inaccuracies, recruitment difficulties, patient adherence issues, etc. Addressing these challenges in trial design can lead to development of more effective study protocols with optimized patient safety and trial efficiency. In this paper, we present a novel method to identify similar historical trials by summarizing clinical trial protocols and searching for similar trials based on a query trial's protocol. Our approach significantly outperforms all baselines, achieving up to a 78\% improvement in recall@1 and a 53\% improvement in precision@1 over the best baseline. We also show that our method outperforms all other baselines in partial trial similarity search and zero-shot patient-trial matching, highlighting its superior utility in these tasks.

\end{abstract}

\section{Introduction} \label{sec:intro}

Clinical trials are vital for advancing medical interventions. However, the success of these trials is largely influenced by the quality of trial design and  risk mitigation strategies \cite{fogel2018factors}. To improve the probability of trial success, similar historical trials are used as references to inform the design of future trials \cite{luo2024clinical}. Historical trials can be used to determine and optimize trial design including aspects like target population, eligibility criteria, mitigation strategies, dosage schedules, and anticipation of risks and adverse events. Identifying similar trials is not a trivial task and requires investigators to search and review numerous historical protocols ---a process that is labor-intensive and error-prone, often involving the manual examination of thousands of studies \cite{luo2024clinical}. Given the importance of historical clinical trials in optimizing trial protocols \cite{wang2022artificial}, it is essential to develop faster, streamlined, and more efficient trial search methods.

While advancements in data mining have improved the efficiency of similar clinical trial retrieval, most efforts have focused largely on section-level retrieval rather than comprehensive protocol-to-protocol matching \cite{roy2019towards, rybinski2021science2cure}. Trial2Vec introduced an initial clinical trial search framework for unsupervised trial similarity search \cite{wang2022trial2vec}. GTSLNet is a recent supervised approach trained on a private dataset of clinical trials labeled by experts \cite{luo2024clinical}. The main sections of a sample clinical trial protocol are listed in Table~\ref{tab:example_trial}. In this paper, we focus solely on clinical trial protocols and refer to them as documents and use the terms ``document'' and ``protocol'' interchangeably.

\begin{table}[!ht]
\resizebox{\columnwidth}{!}{%
\begin{tabular}{|l|l|}
\hline
Title                                                          & \begin{tabular}[c]{@{}l@{}}STunning in Acute Myocardial Infarction\\  - BAS (STAMI-BAS)\end{tabular}                                                                                                                                                     \\ \hline
Description                                                    & \begin{tabular}[c]{@{}l@{}}The objective of this trial is to examine the \\ effect of immediate versus late administration ...\end{tabular}                                                                                              \\ \hline
\begin{tabular}[c]{@{}l@{}}Eligibility\\ Criteria\end{tabular} & \begin{tabular}[c]{@{}l@{}}Inclusion Criteria:\\ 1. Patients with STEMI who undergo primary PCI... \\ \\ 2. Informed consent\\ \\ Exclusion Criteria:\\ 1. Killip class $\geq$ 3\\ 2.Chronic kidney disease with GFR \\\textless 25 ml/min/1.73 m2\end{tabular} \\ \hline
\begin{tabular}[c]{@{}l@{}}Outcome\end{tabular}     & Measurement of Global longitudinal strain (GLS, \%)                                                                                                                                                                                                      \\ \hline
Disease                                                        & Myocardial Infarction                                                                                                                                                                                                                                    \\ \hline
Intervention                                                   & \begin{tabular}[c]{@{}l@{}}1. Bisoprolol Oral Tablet\\ 2. Ramipril Oral Product\\ 3. Dapagliflozin Oral Product\end{tabular}                                                                                                                             \\ \hline
...                                                            & ...                                                                                                                                                                                                                                                      \\ \hline
\end{tabular}}
\caption{An example of clinical trial document drawn from \textit{ClinicalTrials.gov}.}
\label{tab:example_trial}
\vspace{-1em}
\end{table}

The main challenges of developing a method for clinical trial search are: 

\begin{itemize}[left=0pt]
    \item \textbf{Challenge 1: Lack of publicly available labeled data} \label{challenge1} - A significant challenge is the lack of publicly available labeled data needed to train supervised methods. GTSLNet \cite{luo2024clinical} improves over Trial2Vec \cite{wang2022trial2vec} on their private labeled dataset indicating the need for supervised approaches to improve accuracy and effectiveness. 
    \item \textbf{Challenge 2: Lengthy documents} \label{challenge2} - As trial documents often exceed 1,000 words \cite{wang2022trial2vec}, encoding long trial documents by truncating or averaging the embeddings inevitably results in poor retrieval quality. Although Trial2Vec aims to solve the long document problem by encoding different sections of the document separately, some sections of these documents (e.g., \textit{eligibility criteria, description}, etc.) can be larger than the context length of the encoder model used in Trial2Vec leading to truncation of potentially important information in these sections. Moreover, Trial2Vec combines \textit{ eligibility criteria,  description}, etc. long sections into a combined section called \textit{context} instead of separately encoding them. This highlights the need for a clinical trial search method that can address the long document problem (which persists in existing methods including Trial2Vec) while preventing loss of critical information.  
    \item \textbf{Challenge 3: Lack of understanding of local context} \label{challenge3} - Two medical texts can have significant word-wise overlap while describing entirely different concepts, posing a challenge for similarity computation. Trial2Vec \cite{wang2022trial2vec} aims to solve this by extracting medical entities using local contrastive learning. However, two sentences can have exactly the same medical entities but have very different meanings. For example: \textit{``The patient was tested for \textbf{insulin} levels to diagnose their \textbf{diabetes}.''} and 
    \textit{``The patient was prescribed \textbf{insulin} to manage their \textbf{diabetes}.''} both have the same medical entities \textbf{\textit{insulin}} and \textbf{\textit{diabetes}} but have totally different meanings. Better approaches are needed to ensure that the encoder model has an improved understanding of the local context. 
    \item \textbf{Challenge 4: Inefficient contrastive supervision} \label{challenge4} - Though existing methods like SimCSE and Trial2Vec provide contrastive supervision mechanisms to train models with the ability to differentiate between similar (positive) and non-similar (negative) trials, the methodologies used to define `similar' and `non-similar' leave room for improvement. Unsupervised methods such as SimCSE \cite{gao2021simcse} use instance-level contrastive learning by generating positive trial document pairs (i.e., `entailments') by using the same trial document input twice and treating all other trial document inputs as negatives (i.e., `contradictions'). Trial2Vec \cite{wang2022trial2vec} creates positive trial document pairs by omitting sections from the trial document (see different sections in Table 1). However, this approach may result in the loss of critical information across these pairs, causing the model to identify similar trials without accounting for the missing details. 
\end{itemize}

We developed \method, a \texttt{SE}mi-supervised \texttt{C}linical t\texttt{R}ial protocol similari\texttt{T}y searching method, to address the above-mentioned challenges. Our approach minimizes the reliance on very large publicly available labeled datasets (Challenge 1) by using labeled trial similarity data from \cite{wang2025leads} and publicly available unlabeled  data\footnote{\url{https://clinicaltrials.gov}} in a semi-supervised manner. To address the long document problem (Challenge 2), we represent a clinical trial as a set of question-answer (Q/A) pairs (generated by humans and LLMs), which significantly reduces the length of trial documents. To better capture local semantic context (Challenge 3), we train our model contrastively at the Q/A level instead of the entity level. Since the generated Q/A pairs can vary significantly in meaning depending on the original sentences they were derived from, contrastive training in \method ensures that sentences containing the same medical entities but different semantic meanings are assigned distinct embeddings. To tackle inefficient contrastive supervision (Challenge 4), we employ a two-level contrastive approach:
\begin{enumerate}[left=0pt]
\item \textbf{Local (Q/A Level):} Contrastive training at the Q/A level ensures that the model accurately captures local context. Positive samples for each Q/A pair are automatically selected (details in Section \ref{section:Method}).
\vspace{-1pt}
\item \textbf{Global (Trial Level):} Contrastive training at the trial level ensures the model embeds similar trials close together and dissimilar trials far apart in the embedding space.\blockcomment{Positive samples are generated by removing a single Q/A pair from a large section for unlabeled data, while similar trials from labeled data acre used as positive samples.} Positive samples are generated by removing a single Q/A pair from a large section for unlabeled data, and by using similar trials from labeled data. Both hard and soft negatives are utilized to further improve performance. In contrast to existing approaches that use the same trial or drop sections from the trial document to label positive samples (methods that may be too restrictive or may lose critical information needed for trial similarity), our approach minimizes the loss of critical information by summarizing in Q/A pairs while adding flexibility to the search.
\end{enumerate}

This paper compares \method to baseline methods, showing it outperforms them in trial document similarity search while using less than a quarter of the training data required by Trial2Vec. It also achieved superior scores in query-to-trial matching and zero-shot patient-to-trial matching tasks. \blockcomment{The trial embeddings generated by our method can also be used in other downstream tasks (e.g. prediction of trial outcomes/other metrics).} The paper discusses (i) related methods for clinical trial search in Section \ref{section:Related}, (ii) detailed method details of \method in Section \ref{section:Method}, (iii) experimental setting and results in Section \ref{section:Exp}, (iv) conclusion in Section \ref{section:Conclusion}, and (v) limitations and potential future directions in Section \ref{section: Limitation}.

\section{Related Work} \label{section:Related}
\subsection{Text and Document Retrieval}
\subsubsection{General Text}
Dense retrieval methods using distributional word representations, such as Word2Vec \cite{mikolov2013efficient} and GloVe \cite{pennington2014glove} became popular due to their superior performance in capturing semantic similarity compared to traditional methods such as TF-IDF \cite{salton1988term}. In contrast, early information retrieval approaches relied heavily on manual feature engineering \cite{trotman2014improvements, yang2017anserini}. The rise of deep learning models, especially contextualized encoders like BERT \cite{devlin2018bert}, has driven significant advancements in neural retrieval methods \cite{van2016learning, dehghani2017neural, yates2021pretrained}. 

In domains like clinical trials where access to labeled data is limited due to cost, privacy, and other reasons \cite{das2023twin, das2024synrl}, zero-shot learning models become a necessity. Although some approaches have attempted to improve retrieval quality by performing post-processing on pre-trained BERT embeddings \cite{li2021selfdoc}, their performance remains suboptimal without domain-specific training. While BERT-like models fine-tuned or pre-trained on biomedical documents and electronic health records, such as BioBERT \cite{lee2020biobert} and Bio\_ClinicalBERT \cite{alsentzer2019publicly} exist, they are not trained for clinical trial retrieval, resulting in suboptimal performance.

\begin{figure*}[ht]
    \centering
    \includegraphics[width=0.85\textwidth]{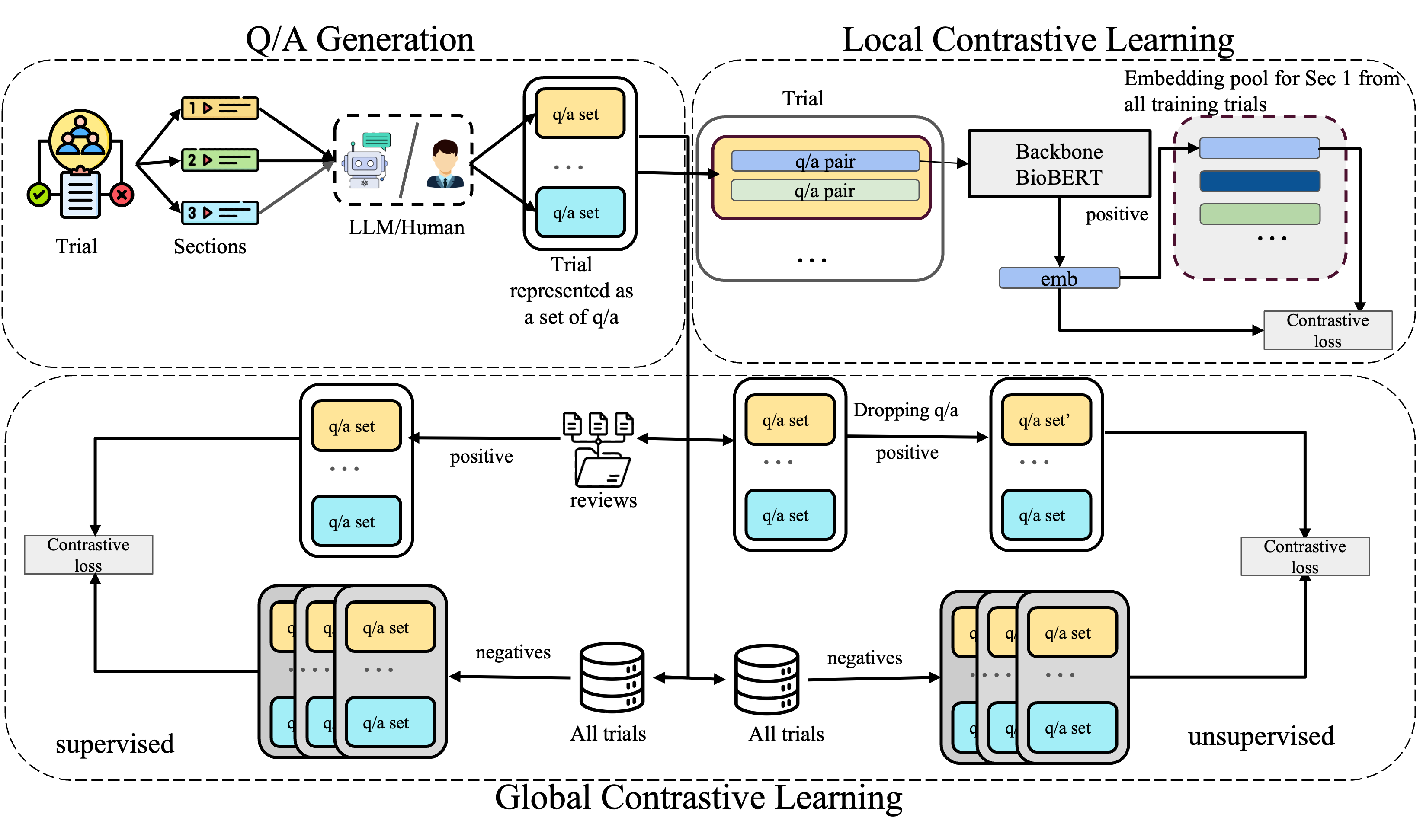}
    \vspace{-1em}
    \caption{Overview of \method. \method consists of three main components: Q/A Generation, Local Contrastive Learning and Global Contrastive Learning. Without loss of generality, we illustrate Sec 1 in the Local Contrastive Learning block, applicable to all sections.}
    \label{fig:pipeline}
    \vspace{-1em}
\end{figure*}

\subsubsection{Clinical Trial}
Traditional clinical trial query engines use rule-based entity matching on trial metadata which relies heavily on databases hence limiting their flexibility \cite{tasneem2012database, tsatsaronis2012ponte, jiang2014cross, park2020interactive}. Recent approaches utilize supervised neural ranking to match trial titles or relevant segments with user queries \cite{roy2019towards, rybinski2021science2cure}. However, these methods are limited to specific parts of the trial documents. PRISM \cite{gupta2024prism} is a recent patient-to-trial matching model that transforms the trial criteria from clinicaltrials.gov into simplified, independent questions, each with answers like "Yes," "No," or "NA." However, we utilize an LLM to generate Q/A pairs from eligibility criteria by extracting key information and the answers are not limited to "Yes," "No," or "NA." Trial2Vec \cite{wang2022trial2vec} is a self-supervised method that encodes entire trial documents, enabling searches based on trial-level similarity and query-based searches. GTSLNet \cite{luo2024clinical}, a supervised approach designed to identify similarity at the trial level, outperforms Trial2Vec and other unsupervised and self-supervised approaches in retrieval tasks, but requires large labeled datasets to avoid overfitting \cite{althnian2021impact}. Self-supervised methods avoid manual annotation by using unlabeled data, though they typically deliver lower performance \cite{luo2024clinical}. To our knowledge, \method introduces the first semi-supervised approach for trial retrieval that balances between the drawbacks of supervised and unsupervised methods and shows improved performance in trial-level retrieval scores. \blockcomment{\method also outperformed baseline methods in  partial trial similarity search and zero-shot patient-to-trial similarity search.}

\subsection{Text Contrastive Learning}
Contrastive learning has recently become a widely discussed topic \cite{chen2020simple, chen2021exploring, carlsson2021semantic, gao2021simcse, aberdam2021sequence, yang2022unified, zhang2020unsupervised, wang2020cross}. This technique has been used for zero-shot retrieval capabilities \cite{wang2020cross, zhang2020unsupervised} and to enhance downstream NLP tasks, such as text classification \cite{pan2022improved, chen2022contrastnet}. However, current approaches focus on improving sentence embeddings by manipulating text alone, making them less effective for handling long clinical trial-specific documents. Trial2Vec  addresses this limitation by generating document embeddings, providing a better solution for such scenarios. 

\section{Proposed Method} \label{section:Method}
\vspace{-0.5em}
In this section, we detail the architecture of \method (Figure \ref{fig:pipeline}). \method is built on three main components: (1) Q/A generation, (2) Local contrastive learning, and (3) Global contrastive learning. First, we generate key Q/A pairs for each trial using LLM. Next, we fine-tune the BioBERT \cite{lee2020biobert} backbone encoder using local and global contrastive learning to generate the final embeddings.

\subsection{Q/A Generation}
We represent a clinical trial with a set of key Q/A pairs generated from different sections of a clinical trial document. The method assumes that two similar documents will have a similar set of key Q/A pairs. This helps reduce the length of the trial documents and better capture the local context (see \ref{sec:local}). We utilize the Llama-3.1-8B-Instruct model to generate Q/A pairs from large sections (e.g. \textit{eligibility criteria}, etc.) that can follow user instructions effectively. For smaller sections (e.g., \textit{title, disease, intervention}, etc.), we use predefined questions generated by humans. An example of a clinical trial represented by a set of Q/A pairs (Figure \ref{tab:chickpea_pulao_study}) and the prompt used to generate them can be found in the Appendix \ref{sec:detail_q_a}. 

\vspace{-0.5em}
\subsection{Local Contrastive Learning} \label{sec:local}

To enhance \method's discriminative power and improve its understanding of local context, we perform contrastive training at the Q/A level. Previous methods such as Trial2Vec rely on entity similarity to compare sentences or documents which could lead to incorrect assumptions when sentences share entities but differ in meaning (Challenge 3). By focusing on Q/A pairs, \method effectively captures these nuances. This also helps in partial trial matching to find the best matching trial documents given the \textit{title} of the query trial (see results in Section \ref{subsection: partial}). Similar to Trial2Vec, BioBERT is used as the backbone encoder for \method.

We finetune the BioBERT backbone at the Q/A level, where positive and negative samples are automatically selected from the training pool. Given a training pair \( (Q_i, A_i) \), the most similar pair \( (Q_p, A_p) \) within the same section's Q/A pool is selected as the positive sample. Let \( v_i \), \( v_i^+ \) denote the embeddings from the backbone (before finetuning) of the anchor and positive Q/A pairs, respectively.
\begin{align} v_i^+ &= \underset{v_j \neq v_i}{\text{argmax}} , \psi(v_i, v_j), \quad v_j, v_i \in \mathcal{P}, \end{align} 
where $\mathcal{P}$ represents the pool of Q/A pairs from the same section as $v_i$ across all trials. 
A similarity function \( \psi(x, y) \) computes the cosine similarity between the embeddings of the Q/A pairs. All other pairs in the batch, excluding the selected positive pair, serve as negative samples. InfoNCE loss for local contrastive learning is defined as:
\begin{align}
\mathcal{L}_{\text{l}} = - \frac{1}{N} \sum_{i=1}^N \log \frac{
    \exp\left(\psi (v_i\text{,} v_i^+) /\tau \right)}
    {\sum_{j=1}^N \exp\left(\psi(v_i \text{,} v_j) / \tau\right)}.
\end{align}

In the equation, $v_i$ and $v_i^+$ represent the embedding for the $i$-th anchor Q/A pair in the batch and the embedding for its corresponding positive Q/A pair associated with the query $v_i$, respectively. $v_j$ refers to all Q/A pairs in the batch. $\tau$ refers to the temperature scaling parameter. $N$ represents number of Q/A pairs in the batch. An example (anchor, positive) pair:
\begin{mdframed}
\begin{verbatim}
('What is the age range for eligible
children? 6-12 years',
'What is the age range for enrollment?
6-12 years')
\end{verbatim}
\end{mdframed}
Here, the anchor Q/A pair is from trial NCT00000113 and the positive Q/A pair is from trial NCT00006565.

\subsection{Global Contrastive Learning} \label{section:Global Contrastive Training}

In global contrastive learning, each trial is represented as a set of Q/A pairs, where the entire set is used as input. The objective of global contrastive training is to learn trial-level embeddings such that positive trials with similar patterns in their Q/A pairs (e.g., similar trials or modified versions of the same trial) are pulled closer in the embedding space and negative trials (e.g., unrelated or hard negatives) with dissimilar patterns are pushed apart.

We utilize both labeled and unlabeled trials for contrastive training. For unlabeled trials, a positive sample for an anchor trial \(T_i\) is created by dropping one Q/A pair from a section with multiple Q/A pairs of that trial, resulting in \(T_i^+\). Hard negatives are chosen as trials that share the same disease indication (similar approach as Trial2Vec) as \(T_i\) in the entire dataset, while other trials in the batch serve as additional negatives. A negative trial for anchor \(T_i\) is denoted as \(T_i^-\). For labeled trials, the positive trial \(T_i^+\) is explicitly selected based on ground truth (provided label), representing a known similar trial. Hard negatives and other negatives are constructed as in the unsupervised setting.
\begin{align}\label{eq:paired}
    \mathcal{L}_{\text{paired}} = &\ - \frac{1}{N} \sum_{i=1}^{N} \log \frac{\exp\left(\psi(z_i\text{,} z_i^+) / \tau \right)}{Z_{\text{p}}}, \\
    Z_{\text{P}} = &\ \exp\left(\psi(z_i\text{,} z_i^+)/ \tau \right) + \exp\left(\psi(z_i\text{,} z_i^-) / \tau \right), \nonumber
\end{align}
\vspace{-1em}
\begin{align}\label{eq:in_batch}
    \mathcal{L}_{\text{in-batch}} = &\ - \frac{1}{N} \sum_{i=1}^{N} \log \frac{
    \exp\left(\psi(z_i\text{,} z_i^+) / \tau\right)}
    {\sum_{j=1}^{N} \exp\left(\psi(z_i\text{,} z_j) / \tau\right)}, 
\end{align}
\vspace{-1em}
\begin{equation}
    \mathcal{L}_{\text{g}} = \mathcal{L}_{\text{paired}} + \mathcal{L}_{\text{in-batch}}.
\end{equation}

In the equations for global contrastive learning, the following notations are used. $z_i$ represents the query trial embedding for the $i$-th trial, and $z_i^+$ is its positive trial's embedding. In the case of the paired loss (Eq. \ref{eq:paired}), $z_i^-$ refers to an explicit negative trial embedding corresponding to $z_i$. $\tau$ is the temperature scaling parameter.  $Z_{\text{p}}$ is the normalization term. For in-batch loss (Eq. \ref{eq:in_batch}), the denominator is the sum over all trials in the batch, where $z_j$ represents all trial embeddings in the batch. $N$ denotes batch size. \( \psi() \) computes cosine similarity.

Finally, we use cosine similarity on the trial embeddings to rank clinical trials given a query trial.

\begin{table*}[ht]
\centering
\resizebox{\textwidth}{!}{%
\begin{tabular}{lcccccccc}
\toprule
 & \textbf{precision@1} & \textbf{recall@1} & \textbf{precision@2} & \textbf{recall@2} & \textbf{precision@5} & \textbf{recall@5} & \textbf{nDCG@5} & \textbf{MAP} \\
\midrule
TF-IDF & 0.363 $\pm$ 0.073 & 0.244 $\pm$ 0.054 & 0.298 $\pm$ 0.048 & 0.388 $\pm$ 0.064 & 0.217 $\pm$ 0.023 & 0.687 $\pm$ 0.067 & 0.522 $\pm$ 0.055 & 0.501 $\pm$ 0.050 \\
BM25 & 0.334 $\pm$ 0.071 & 0.223 $\pm$ 0.055 & 0.271 $\pm$ 0.050 & 0.350 $\pm$ 0.066 & 0.180 $\pm$ 0.026 & 0.567 $\pm$ 0.079 & 0.454 $\pm$ 0.065 & 0.471 $\pm$ 0.052 \\
Word2Vec & 0.293 $\pm$ 0.071 & 0.184 $\pm$ 0.047 & 0.266 $\pm$ 0.049 & 0.328 $\pm$ 0.061 & 0.191 $\pm$ 0.024 & 0.573 $\pm$ 0.067 & 0.435 $\pm$ 0.058 & 0.435 $\pm$ 0.049 \\
BERT & 0.241 $\pm$ 0.067 & 0.130 $\pm$ 0.040 & 0.235 $\pm$ 0.046 & 0.279 $\pm$ 0.064 & 0.197 $\pm$ 0.024 & 0.591 $\pm$ 0.064 & 0.415 $\pm$ 0.052 & 0.400 $\pm$ 0.043 \\
BioBERT & 0.347 $\pm$ 0.078 & 0.202 $\pm$ 0.053 & 0.275 $\pm$ 0.051 & 0.313 $\pm$ 0.068 & 0.202 $\pm$ 0.022 & 0.612 $\pm$ 0.061 & 0.462 $\pm$ 0.057 & 0.450 $\pm$ 0.051 \\
Bio\_ClinicalBERT & 0.280 $\pm$ 0.071 & 0.169 $\pm$ 0.048 & 0.261 $\pm$ 0.050 & 0.317 $\pm$ 0.054 & 0.207 $\pm$ 0.024 & 0.644 $\pm$ 0.058 & 0.464 $\pm$ 0.050 & 0.437 $\pm$ 0.046 \\
Longformer & 0.253 $\pm$ 0.070 & 0.158 $\pm$ 0.049 & 0.272 $\pm$ 0.050 & 0.338 $\pm$ 0.065 & 0.198 $\pm$ 0.025 & 0.631 $\pm$ 0.067 & 0.447 $\pm$ 0.056 & 0.425 $\pm$ 0.048 \\
Clinical-Longformer & 0.206 $\pm$ 0.065 & 0.123 $\pm$ 0.042 & 0.211 $\pm$ 0.050 & 0.252 $\pm$ 0.066 & 0.168 $\pm$ 0.021 & 0.533 $\pm$ 0.065 & 0.369 $\pm$ 0.051 & 0.369 $\pm$ 0.045 \\
IDCM & 0.156 $\pm$ 0.054 & 0.102 $\pm$ 0.040 & 0.167 $\pm$ 0.039 & 0.206 $\pm$ 0.056 & 0.132 $\pm$ 0.019 & 0.391 $\pm$ 0.059 & 0.286 $\pm$ 0.048 & 0.324 $\pm$ 0.039 \\
Trial2Vec & 0.422 $\pm$ 0.078 & 0.263 $\pm$ 0.054 & 0.375 $\pm$ 0.060 & 0.458 $\pm$ 0.068 & 0.227 $\pm$ 0.029 & 0.689 $\pm$ 0.067 & 0.553 $\pm$ 0.064 & 0.539 $\pm$ 0.058 \\
\method & \textbf{0.647 $\pm$ 0.077} & \textbf{0.467 $\pm$ 0.063} & \textbf{0.508 $\pm$ 0.046} & \textbf{0.682 $\pm$ 0.061} & \textbf{0.297 $\pm$ 0.023} & \textbf{0.924 $\pm$ 0.034} & \textbf{0.796 $\pm$ 0.042} & \textbf{0.754 $\pm$ 0.044} \\

\bottomrule
\end{tabular}}
\caption{Performance evaluation of retrieval models for complete trial similarity search on the labeled test set. The table presents precision, recall, nDCG, and MAP metrics, reported as mean $\pm$ standard deviation, with the best values highlighted in \textbf{bold}.}

\label{table:protocol2protocol}
\end{table*}

\begin{table*}[!ht]
\centering
\resizebox{\textwidth}{!}{%
\begin{tabular}{lcccccccc}
\toprule
 & \textbf{precision@1} & \textbf{recall@1} & \textbf{precision@2} & \textbf{recall@2} & \textbf{precision@5} & \textbf{recall@5} & \textbf{nDCG@5} & \textbf{MAP} \\
\midrule
TF-IDF & 0.359 $\pm$ 0.078 & 0.252 $\pm$ 0.058 & 0.320 $\pm$ 0.055 & 0.410 $\pm$ 0.062 & 0.214 $\pm$ 0.023 & 0.664 $\pm$ 0.054 & 0.517 $\pm$ 0.054 & 0.505 $\pm$ 0.052 \\
BM25 & 0.363 $\pm$ 0.074 & 0.248 $\pm$ 0.052 & 0.298 $\pm$ 0.050 & 0.390 $\pm$ 0.060 & 0.199 $\pm$ 0.024 & 0.627 $\pm$ 0.064 & 0.493 $\pm$ 0.053 & 0.491 $\pm$ 0.048 \\
Word2Vec & 0.308 $\pm$ 0.079 & 0.203 $\pm$ 0.055 & 0.237 $\pm$ 0.050 & 0.298 $\pm$ 0.065 & 0.190 $\pm$ 0.024 & 0.595 $\pm$ 0.072 & 0.447 $\pm$ 0.062 & 0.442 $\pm$ 0.053 \\
BERT & 0.233 $\pm$ 0.057 & 0.137 $\pm$ 0.037 & 0.204 $\pm$ 0.042 & 0.241 $\pm$ 0.048 & 0.179 $\pm$ 0.022 & 0.565 $\pm$ 0.064 & 0.392 $\pm$ 0.046 & 0.385 $\pm$ 0.036 \\
BioBERT & 0.288 $\pm$ 0.078 & 0.191 $\pm$ 0.059 & 0.244 $\pm$ 0.055 & 0.318 $\pm$ 0.079 & 0.192 $\pm$ 0.024 & 0.601 $\pm$ 0.074 & 0.444 $\pm$ 0.065 & 0.439 $\pm$ 0.056 \\
Bio\_ClinicalBERT & 0.257 $\pm$ 0.083 & 0.172 $\pm$ 0.063 & 0.203 $\pm$ 0.047 & 0.276 $\pm$ 0.069 & 0.180 $\pm$ 0.023 & 0.578 $\pm$ 0.069 & 0.416 $\pm$ 0.059 & 0.413 $\pm$ 0.051 \\
Longformer & 0.235 $\pm$ 0.067 & 0.163 $\pm$ 0.049 & 0.199 $\pm$ 0.046 & 0.264 $\pm$ 0.059 & 0.156 $\pm$ 0.022 & 0.490 $\pm$ 0.067 & 0.364 $\pm$ 0.054 & 0.385 $\pm$ 0.044 \\
Clinical-Longformer & 0.238 $\pm$ 0.066 & 0.160 $\pm$ 0.045 & 0.221 $\pm$ 0.045 & 0.303 $\pm$ 0.056 & 0.181 $\pm$ 0.025 & 0.573 $\pm$ 0.063 & 0.413 $\pm$ 0.050 & 0.412 $\pm$ 0.041 \\
IDCM & 0.306 $\pm$ 0.073 & 0.213 $\pm$ 0.055 & 0.273 $\pm$ 0.049 & 0.369 $\pm$ 0.070 & 0.180 $\pm$ 0.024 & 0.584 $\pm$ 0.079 & 0.452 $\pm$ 0.064 & 0.462 $\pm$ 0.051 \\
Trial2Vec & 0.456 $\pm$ 0.088 & 0.322 $\pm$ 0.065 & 0.370 $\pm$ 0.057 & 0.482 $\pm$ 0.068 & 0.228 $\pm$ 0.027 & 0.717 $\pm$ 0.066 & 0.592 $\pm$ 0.062 & 0.579 $\pm$ 0.056 \\
\method & \textbf{0.548 $\pm$ 0.083} & \textbf{0.390 $\pm$ 0.066} & \textbf{0.465 $\pm$ 0.044} & \textbf{0.623 $\pm$ 0.059} & \textbf{0.289 $\pm$ 0.023} & \textbf{0.902 $\pm$ 0.040} & \textbf{0.745 $\pm$ 0.044} & \textbf{0.696 $\pm$ 0.047} \\
\bottomrule
\end{tabular}}
\caption{Performance evaluation of retrieval models for query-to-trial matching (partial trial similarity search) on the labeled test set. Metrics include precision, recall, nDCG, and MAP, reported as mean $\pm$ standard deviation. Best-performing results are highlighted in \textbf{bold}.}
\label{tab:query2trial}
\vspace{-1em}
\end{table*}

\begin{table*}[]
\resizebox{\textwidth}{!}{%
\begin{tabular}{@{}lllllllll@{}}
\toprule
                & \textbf{precision@1} & \textbf{recall@1} & \textbf{precision@2} & \textbf{recall@2} & \textbf{precision@5} & \textbf{recall@5} & \textbf{nDCG@5} & \textbf{MAP} \\ \midrule
TF-IDF & 0.494 $\pm$ 0.064 & 0.097 $\pm$ 0.012 & 0.529 $\pm$ 0.043 & 0.217 $\pm$ 0.020 & 0.536 $\pm$ 0.029 & 0.546 $\pm$ 0.033 & 0.541 $\pm$ 0.030 & 0.641 $\pm$ 0.023 \\
BM25 & 0.527 $\pm$ 0.065 & 0.112 $\pm$ 0.016 & 0.530 $\pm$ 0.043 & 0.218 $\pm$ 0.019 & 0.493 $\pm$ 0.023 & 0.504 $\pm$ 0.023 & 0.514 $\pm$ 0.025 & 0.623 $\pm$ 0.020 \\
Word2Vec & 0.523 $\pm$ 0.070 & 0.102 $\pm$ 0.014 & 0.551 $\pm$ 0.056 & 0.218 $\pm$ 0.022 & 0.546 $\pm$ 0.028 & 0.548 $\pm$ 0.027 & 0.548 $\pm$ 0.032 & 0.642 $\pm$ 0.026 \\
BERT & 0.483 $\pm$ 0.066 & 0.095 $\pm$ 0.013 & 0.519 $\pm$ 0.047 & 0.218 $\pm$ 0.020 & 0.513 $\pm$ 0.032 & 0.520 $\pm$ 0.030 & 0.516 $\pm$ 0.033 & 0.619 $\pm$ 0.026 \\
BioBERT & 0.610 $\pm$ 0.067 & 0.125 $\pm$ 0.015 & 0.574 $\pm$ 0.044 & 0.244 $\pm$ 0.022 & 0.563 $\pm$ 0.022 & 0.582 $\pm$ 0.024 & 0.586 $\pm$ 0.024 & 0.659 $\pm$ 0.019 \\
Bio\_ClinicalBERT & 0.565 $\pm$ 0.074 & 0.122 $\pm$ 0.020 & 0.601 $\pm$ 0.052 & 0.251 $\pm$ 0.024 & 0.552 $\pm$ 0.027 & 0.570 $\pm$ 0.027 & 0.575 $\pm$ 0.030 & 0.659 $\pm$ 0.026 \\
Longformer & 0.534 $\pm$ 0.079 & 0.106 $\pm$ 0.016 & 0.532 $\pm$ 0.049 & 0.210 $\pm$ 0.019 & 0.539 $\pm$ 0.025 & 0.546 $\pm$ 0.024 & 0.545 $\pm$ 0.028 & 0.628 $\pm$ 0.023 \\
Clinical-Longformer & 0.460 $\pm$ 0.069 & 0.095 $\pm$ 0.015 & 0.499 $\pm$ 0.047 & 0.200 $\pm$ 0.019 & 0.479 $\pm$ 0.029 & 0.487 $\pm$ 0.031 & 0.487 $\pm$ 0.031 & 0.598 $\pm$ 0.022 \\
IDCM & 0.353 $\pm$ 0.071 & 0.068 $\pm$ 0.014 & 0.443 $\pm$ 0.052 & 0.178 $\pm$ 0.021 & 0.516 $\pm$ 0.026 & 0.521 $\pm$ 0.024 & 0.493 $\pm$ 0.029 & 0.586 $\pm$ 0.023 \\
Trial2Vec & 0.608 $\pm$ 0.071 & 0.129 $\pm$ 0.019 & 0.598 $\pm$ 0.051 & 0.250 $\pm$ 0.022 & 0.602 $\pm$ 0.031 & 0.616 $\pm$ 0.026 & 0.618 $\pm$ 0.031 & 0.695 $\pm$ 0.025 \\
\method & \textbf{0.710 $\pm$ 0.073} & \textbf{0.158 $\pm$ 0.021} & \textbf{0.682 $\pm$ 0.057} & \textbf{0.292 $\pm$ 0.027} & \textbf{0.627 $\pm$ 0.034} & \textbf{0.641 $\pm$ 0.031} & \textbf{0.666 $\pm$ 0.036} & \textbf{0.744 $\pm$ 0.029} \\
\bottomrule
\end{tabular}}
\caption{Performance evaluation of retrieval models for patient-to-trial matching on a subset of the TREC2021 labeled test set. The table presents precision, recall, nDCG, and MAP metrics, reported as mean $\pm$ standard deviation, with the best values highlighted in \textbf{bold}.}
\label{tab:patient2protocol}
\vspace{-1em}
\end{table*}

\section{Experiments and Results}\label{section:Exp}
\subsection{Datasets and Setup}

To retrieve similar trial documents from full or partial query trials, we trained the model at the global level using around 10,000 labeled trials \cite{wang2025leads} and 60,000 unlabeled trials downloaded from \url{https://aact.ctti-clinicaltrials.org}. This is less than one-fourth of the training data used in Trial2Vec. All datasets are in English. For each review paper utilized in \cite{wang2025leads}, a set of trials is identified as similar based on shared characteristics, such as diseases, interventions, population and outcomes, sourced from Cochrane Reviews.\footnote{\url{https://www.cochranelibrary.com}} From the remaining labeled trials, we prepared validation and test sets. Each query trial in these sets has 10 corresponding trials, labeled as `relevant' (i.e., positive) or `not relevant' (i.e., negative). Relevant trials were identified from review data, while negative trials were chosen from the same disease category but excluded from the training data and relevant set, with random sampling used if no such trials were available. The test set has 1,420 pairs, and the validation set has 2,000 pairs.

We also ran experiments to evaluate how \method performs zero-shot on patient-to-trial matching. For this task, we used 75 patients from the TREC2021 dataset.\footnote{\url{http://www.trec-cds.org/2021.html}} The dataset has ground-truth labels that indicate each patient's best match to a trial. We prepared a test set where for each patient, there are 10 trials with ground truth labels (both relevant and not relevant trials) resulting in 731 unique test trials for this task. 

We evaluated performance using precision@$k$, recall@$k$, nDCG@$5$ and MAP where $k$ can be 1, 2, or 5. Details about the metrics are available in the Appendix \ref{sec:metrics}. 
\subsection{Implementation Details} We used a large language models (LLM) to generate Q/A for \textit{eligibility criteria} which is the largest section among all sections we use. For all the other sections (e.g., \textit{title, disease, intervention, keywords, outcome}), we use predefined questions. We conducted local contrastive learning for 10 epochs with a batch size of 32. Then, we fine-tuned the model at the trial level for an additional 10 epochs, varying the batch sizes (16 and 32) to identify the configuration that yielded the highest validation scores. The validation results, shown in Figure \ref{fig:precision} in Appendix \ref{sec:appendix}, indicate that a batch size of 16 for \method achieved the best performance. For both local and global contrastive learning, we selected the best model based on validation scores across all epochs. For optimization, we used a learning rate of 2e-5 for local and 1e-6 for global contrastive training, employing the AdamW optimizer \cite{loshchilov2017decoupled}. We leveraged mixed precision during training, which reduced the computational resources required and accelerated the training process. We set \( \tau \) to the default value (0.1) used in the implementation of InfoNCE loss \cite{oord2018representation}. \footnote{\url{https://pypi.org/project/info-nce-pytorch/}} The experiments were carried out using 2 RTX 6000 GPUs. We bootstrapped 50 samples for 100 iterations, then calculated the average score and standard deviation.

\vspace{-0.5em}
\subsection{Baselines}
Due to the lack of ground truth labels for most training samples, we focus on unsupervised and self-supervised baselines for retrieval (similar to Trial2Vec). The baselines include TF-IDF \cite{salton1988term}, BM25 \cite{trotman2014improvements}, Word2Vec \cite{mikolov2013efficient}, BERT \cite{devlin2018bert}, BioBERT \cite{lee2020biobert}, Bio\_ClinicalBERT \cite{alsentzer2019publicly}, Longformer \cite{beltagy2020longformer}, Clinical\_Longformer \cite{li2022clinical}, IDCM \cite{hofstatter2021intra}, and Trial2Vec \cite{wang2022trial2vec}. IDCM, Longformer, and Clinical\_Longformer are designed for long documents, while TF-IDF, BM25, Word2Vec, BERT, and Longformer are general retrieval methods. BioBERT, Bio\_ClinicalBERT, and Clinical\_Longformer are tailored to the biomedical and clinical domains. Among all baselines, only Trial2Vec is specifically trained to retrieve clinical trials based on protocol similarity and we use their precomputed trial embeddings. For \method, we consider a trial to be a set of Q/A pairs from different sections. For other baselines, we concatenate the full text of these sections.

\subsection{Complete Trial Similarity Search}
\vspace{-0.5em}
Given the protocol of a query trial, we evaluate the performance across models to retrieve trials with similar protocols.
As shown in Table \ref{table:protocol2protocol}, \method outperforms all baselines by a significant margin, achieving up to 78\% improvement in recall@1 and 53\% improvement in precision@1 over the best baseline. Improvements are also seen in other metrics, with \method surpassing the best baseline by around 30\%-40\%. The precision and recall gaps between \method and the baselines are larger when $k$ is small. As $k$ increases, precision@$k$ decreases for all methods due to the increased chance of selecting dissimilar trials the more trials are selected. It is important to note that there is a limited number of positive pairs (fewer than 5) relative to the 10 candidate trials (Eq. \ref{eq:5}). Recall@$k$ improves with larger $k$ because it allows retrieval of more relevant items, increasing the proportion of relevant items in the retrieved set. 

\begin{figure*}[h] 
    \centering
    \begin{subfigure}[b]{0.45\linewidth} 
        \centering        \includegraphics[width=\linewidth]{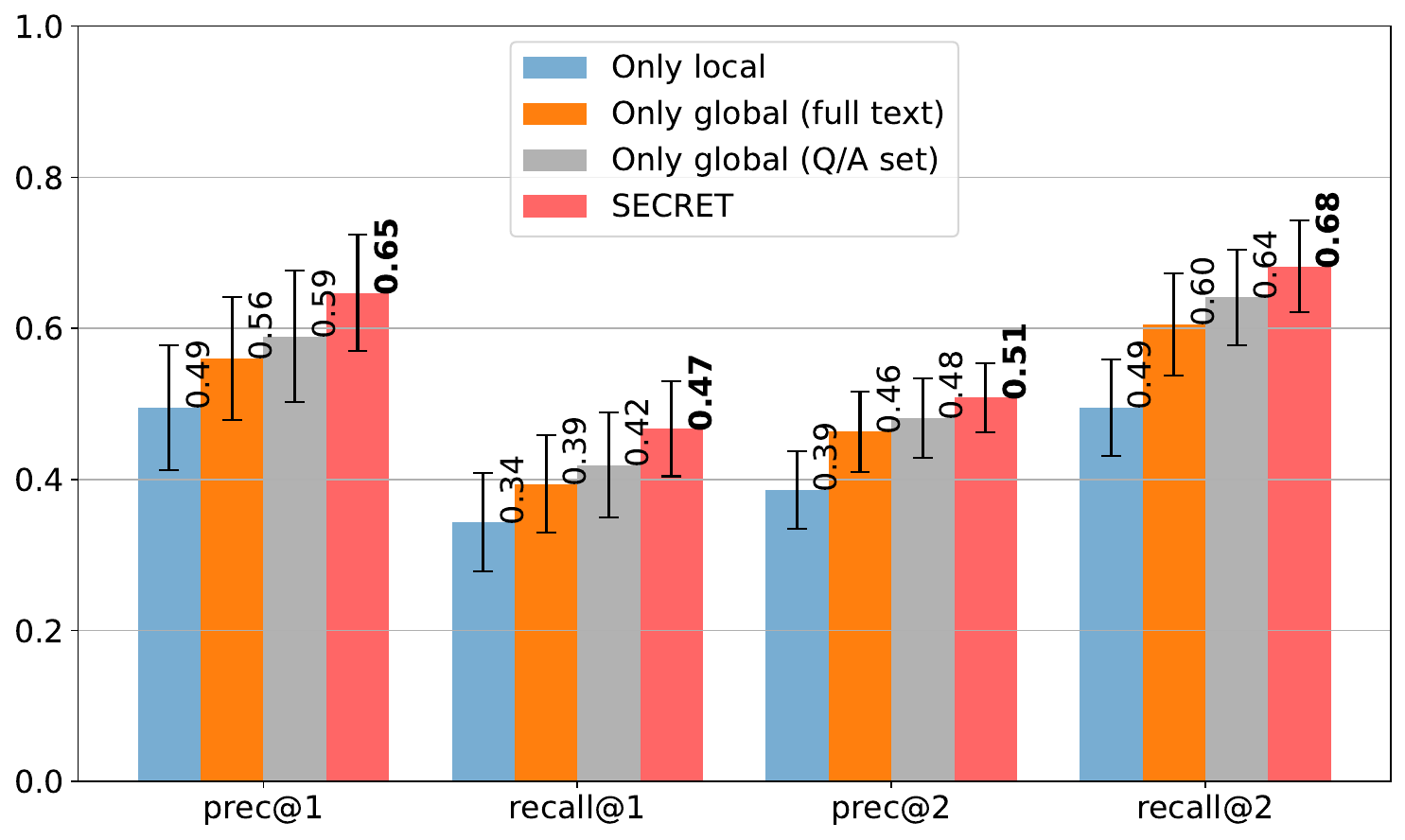}
        \caption{Ablation results by dropping different levels of contrastive learning.}
        \label{fig:ablation}
    \end{subfigure}
    \hfill
    \begin{subfigure}[b]{0.45\linewidth}
        \centering
        \includegraphics[width=\linewidth]{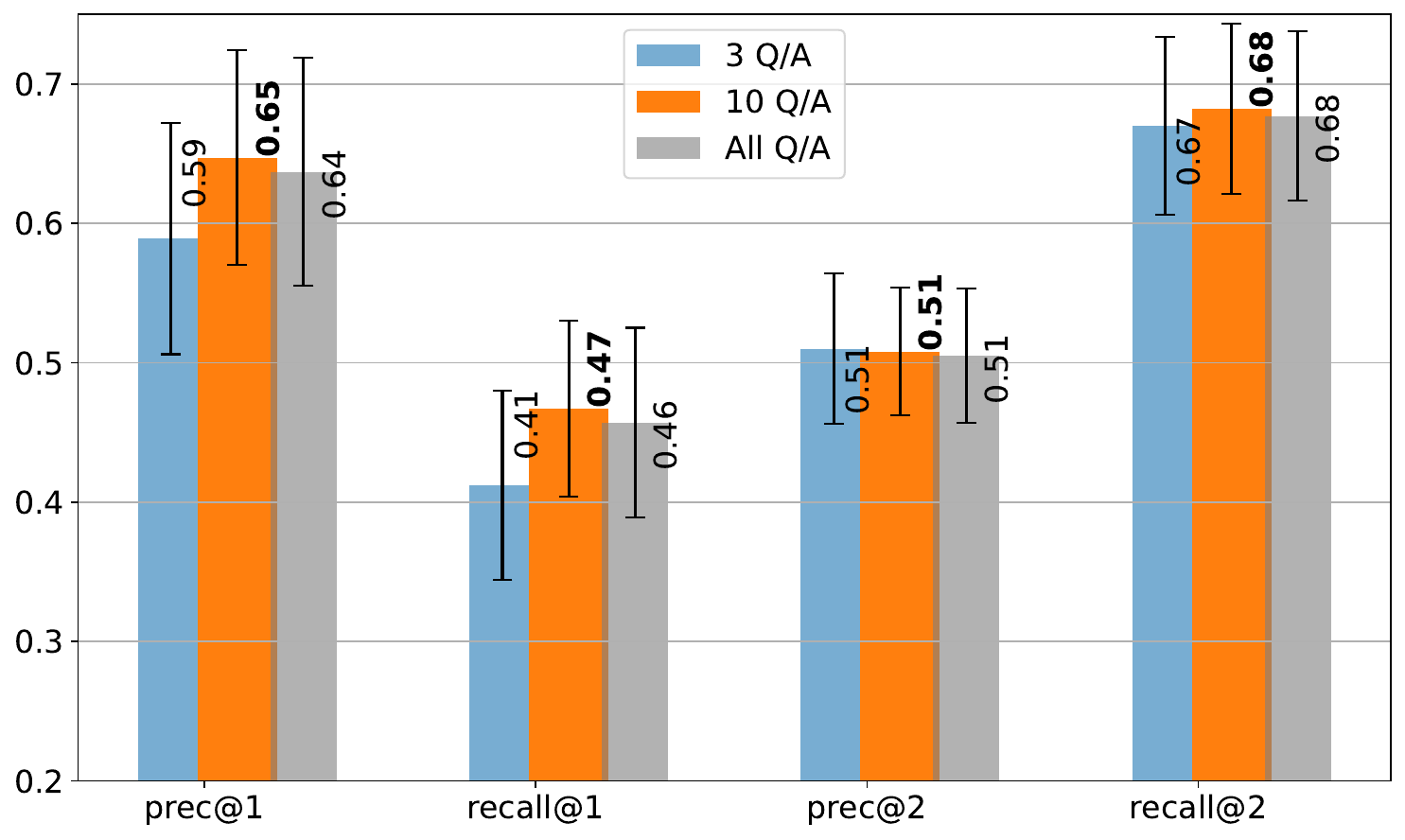}
        \caption{Comparison of performance metrics across different Q/A counts.}
        \label{fig:q_a_length}
    \end{subfigure}
    \label{fig:three_figures}
    \caption{Ablation results}
\end{figure*}

\blockcomment{
\hfill
    \begin{subfigure}[b]{0.32\linewidth}
        \centering
        \includegraphics[width=\linewidth]{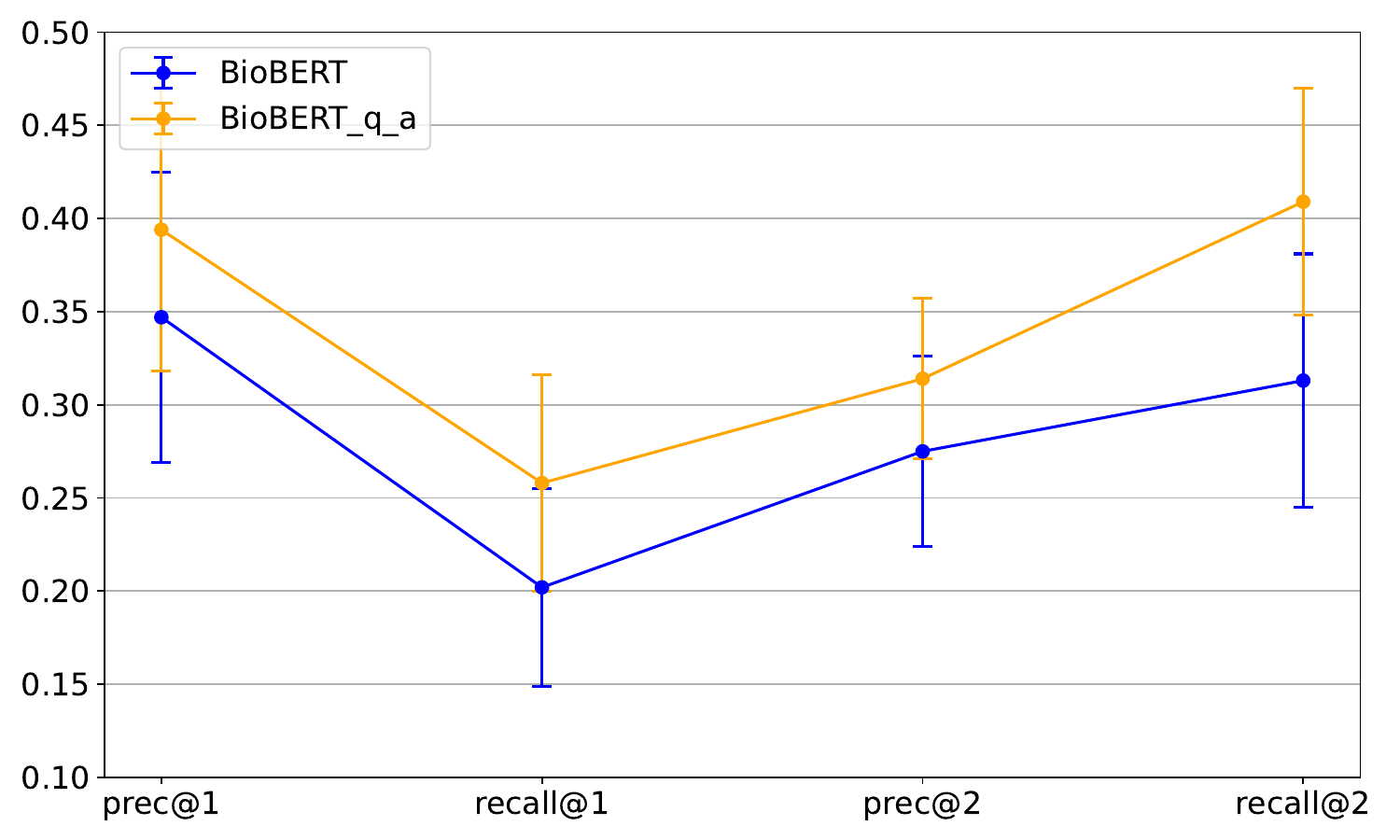}
        \caption{Performance evaluation of BioBERT and BioBERT\_q\_a.}
        \label{fig:biobert_vs_q_a}
    \end{subfigure}
}
\begin{table*}[h!]
\centering
\renewcommand{\arraystretch}{1.2} 
\resizebox{\textwidth}{!}{%
\begin{tabular}{@{}p{5cm}p{4cm}p{4cm}p{4cm}@{}}
\toprule
\textbf{} & \textbf{Query Trial} & \textbf{Trial2Vec Result} & \textbf{\method Result} \\
\midrule
\textbf{NCTID} & NCT00061594 & NCT03470103 & NCT00433017 \\
\textbf{Title} & A Study to Compare rhuFab V2 With Verteporfin ... Macular Degeneration. & A Study in Patients With Wet Age-related Macular Degeneration ... & Verteporfin Photodynamic Therapy ... Age-related Macular Degeneration. \\
\textbf{Intervention} & Ranibizumab & Eylea (Aflibercept, VEGF Trap-Eye, BAY86-5321) & Verteporfin Photodynamic Therapy, Ranibizumab \\
\textbf{Disease} & Macular Degeneration & Macular Degeneration & Macular Degeneration \\
\textbf{Age} & $\geq$50 years & $\geq$18 years & $\geq$50 years \\
\bottomrule
\end{tabular}}
\caption{Case study comparing the retrieval performance of the \method and Trial2Vec.}
\label{tab:case1}
\vspace{-1em}
\end{table*}

\vspace{-1em}
\subsection{Partial Trial Similarity Search} 
\label{subsection: partial}
We evaluated performance across the models on the partial trial retrieval scenario, where users aim to find similar trials based on short or incomplete descriptions (partial attributes). We utilize \textit{title} as a partial attribute. As shown in Table \ref{tab:query2trial}, \method outperforms all baselines by a substantial margin, achieving up to a 29\% improvement in recall@2 over the best baseline. Furthermore, \method shows improvements in other metrics that surpass the best baseline by 20\%-27\%. The evaluation in Figure \ref{fig:partial} in the Appendix \ref{sec:partial_add} shows that combining the title with additional sections consistently improves both precision@1 and recall@1 compared to the title-only approach. Combining the title with intervention achieves the highest scores.

\subsection{Patient-to-trial Similarity Search}
We evaluated the performance of \method in zero-shot patient-to-trial matching, where each patient is represented by a clinical note (patient summary). The patient and trial embeddings are generated using \method and cosine similarity between embeddings is used to rank the trials. We do not train \method for patient-to-trial matching. As shown in Table \ref{tab:patient2protocol}, \method outperforms all baselines. It outperforms the best baseline with improvements of up to 22\% in recall and 17\% in precision.

\subsection{Ablation Studies}
We conducted ablation studies by removing either local or global contrastive training from \method. As shown in Figure \ref{fig:ablation}, the \textit{only local} approach yields the lowest performance, followed by two \textit{only global} approaches, which improve over \textit{only local} approach. Representing trials with Q/A pairs (\textit{only global (Q/A set)}) achieved better scores than using the full text of clinical trial protocol sections (\textit{only global (full text)}). \method, which combines both local and global training, outperforms both methods. Local training focuses on finer, context-specific details, while global training captures broader contextual information. Taking into account both the finer details and the broader context, \method outperforms both individual approaches. A table showing results across all metrics can be found in the Appendix (Table \ref{tab:ablation}).

We have also experimented with the impact of the number of Q/A pairs on retrieval scores. Although the system prompt for the LLM instructed to generate 3-10 question-answer pairs, we observed that the model produced more than 10 pairs at times (see Figure \ref{fig:ques_distribution} in Appendix \ref{sec:appendix}). We selected top 3, 10, and all question-answer pairs generated by the LLM. We then used these selected pairs, along with predefined Q/A pairs for small sections, to separately train our model. Using \textit{`10 Q/A pairs'} achieved the best overall performance, outperforming both the \textit{`all Q/A pairs'} and \textit{`3 Q/A pairs'} settings in precision@1 and recall@1, while achieving comparable scores in precision@2 and recall@2. This indicates the importance of selecting the optimal number of Q/A pairs where too many Q/A may obscure critical information with irrelevant details, while too few Q/A may result in missing essential information (Figure \ref{fig:q_a_length}).

We also evaluated BERT and BioBERT encoders to assess the impact of using Q/A pairs for representing trials, finding consistent improvements over full text (Figures \ref{fig:bert_comparision}, \ref{fig:biobert_comparision} in Appendix \ref{sec:ablation_plus}).

\blockcomment{We also performed experiments with our BioBERT backbone encoder to evaluate the importance of representing trials with Q/A pairs compared to simply using the full text (Figure \ref{fig:biobert_vs_q_a}). Here, BioBERT\_q\_a refers to representing trials using Q/A pairs instead of the full protocol. BioBERT\_q\_a led to a significant and consistent improvement of performance over using the full text of the sections of the clinical trial protocol. This may be due to the truncation of essential information depending on the length of the trial protocol when ingesting entire sections of the trial protocols. More results can be found in Appendix \ref{sec:ablation_plus}.}

\subsection{Case Study}
We conducted a qualitative analysis of the similarity search results (Table \ref{tab:case1}). The top-1 relevant clinical trial retrieved by \method is more closely aligned with the query trial and matches the ground truth label. In contrast, the top-1 trial retrieved by Trial2Vec shares the same disease as the query trial but differs in other details such as interventions and age. Due to the enhanced local context understanding provided by contrastive learning at the question-answer level in \method, the \textbf{Age} attribute (a common eligibility criterion) in the query trial and the top-1 trial retrieved by \method is identical. An additional case study can be found in the Appendix (Table \ref{tab:case2}).

\section{Conclusion} \label{section:Conclusion}
Efficient retrieval of similar clinical trials is critical for optimal design and evaluation of clinical trials, yet the process remains labor-intensive, inefficient and time-consuming. To address this, we developed \method, a semi-supervised clinical trial document similarity search method that reduces reliance on large labeled datasets, resolves the long document problem by a novel representation of trials as question-answer (Q/A) pairs, and captures both local and global semantic contexts through Q/A-level and trial-level contrastive training. Our approach outperforms existing baselines, including Trial2Vec, while requiring significantly less training data. \method achieves superior results in complete trial search, partial trial search, and zero-shot patient-to-trial matching tasks. In summary, \method offers an efficient solution for clinical trial retrieval, setting a new standard in the field of long document retrieval.

\section{Limitations} \label{section: Limitation} 
The current investigation limited evaluations of trial protocols to the following sections: \textit{title, disease, intervention, keywords, outcome} and \textit{eligibility criteria}. We excluded  \textit{description and study design} sections from consideration, which are also lengthy components of the protocol. Although, the exclusion was due to resource constraints in using large language models (LLMs), we hypothesized that the included sections would be sufficient to differentiate between trials. Other related trial documents, such as informed consent forms and adverse event reports, were not included in these experiments. To avoid the complexity of parsing PDFs and the limited accessibility of these documents for many trials, we focused exclusively on trial protocols. Despite these limitations, our approach outperforms the baseline methods. Future work may explore the utility of addition of trial metadata (e.g., forms, fields collected), other related documents related to clinical trials and medical domain knowledge for trial similarity search. We also want to investigate the importance of different sections in the trial search and improve our method on that basis.  

\bibliography{custom}
\appendix
\section{Appendix}
\label{sec:appendix}

\subsection{Details of Q/A Generated by LLM}\label{sec:detail_q_a}
\begin{figure}[!h]
    \centering  \includegraphics[width=0.35\textwidth]{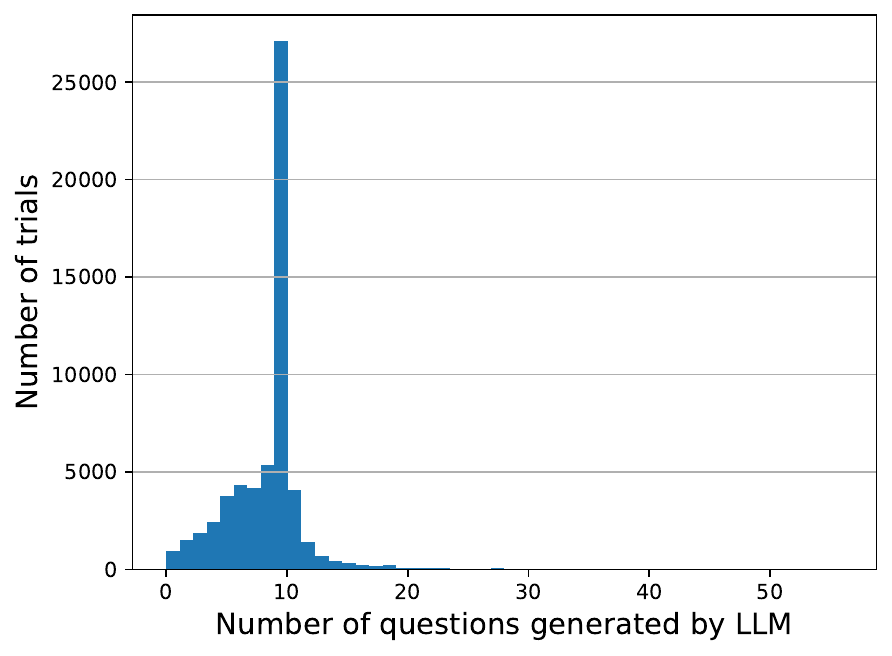}
    \caption{Distribution of Q/A Count per Trial via LLM}
    \label{fig:ques_distribution}
\end{figure}

\blockcomment{The number of Q/A pairs generated for a section varies depending on its length.} We used LLMs to process the \textit{eligibility criteria} section, which is typically a lengthy section (compared to other sections like \textit{title, disease, intervention, outcome,} etc.) and includes both inclusion and exclusion criteria of a trial. The average number of words in the protocol is 312.77 whereas, the average number of words in the set of Q/A pairs is 254.05. Since the length of this section can vary across trials, we prompted the LLM to generate 3-10 Q/A pairs based on the section's length. However, the LLM occasionally generated more than 10 pairs. Figure \ref{fig:ques_distribution} presents a histogram that illustrates the distribution of the number of Q/A pairs generated by the LLM. 

The prompt we used to generate Q/A pairs is as follows:

\textit{``You are an expert at creating key
questions from a medical text and
extracting the answers from the text.
Extract 3-10 Q/A pairs without
repetitions of key entities in the
Q/As. Avoid general questions like
'What are the exclusion criteria?'
Make sure that an answer is no more than
5 tokens/words. Output only json-formated Q/A pairs like this:
\{'Question': 'question1', 'Answer':
'answer1'\} \{'Question': 'question2'
, 'Answer': 'answer2'\} \\ ... \\ Input:"}

\subsection{Hyperparameter Tuning Results}

\begin{figure} [ht]
    \centering    \includegraphics[width=0.5\textwidth]{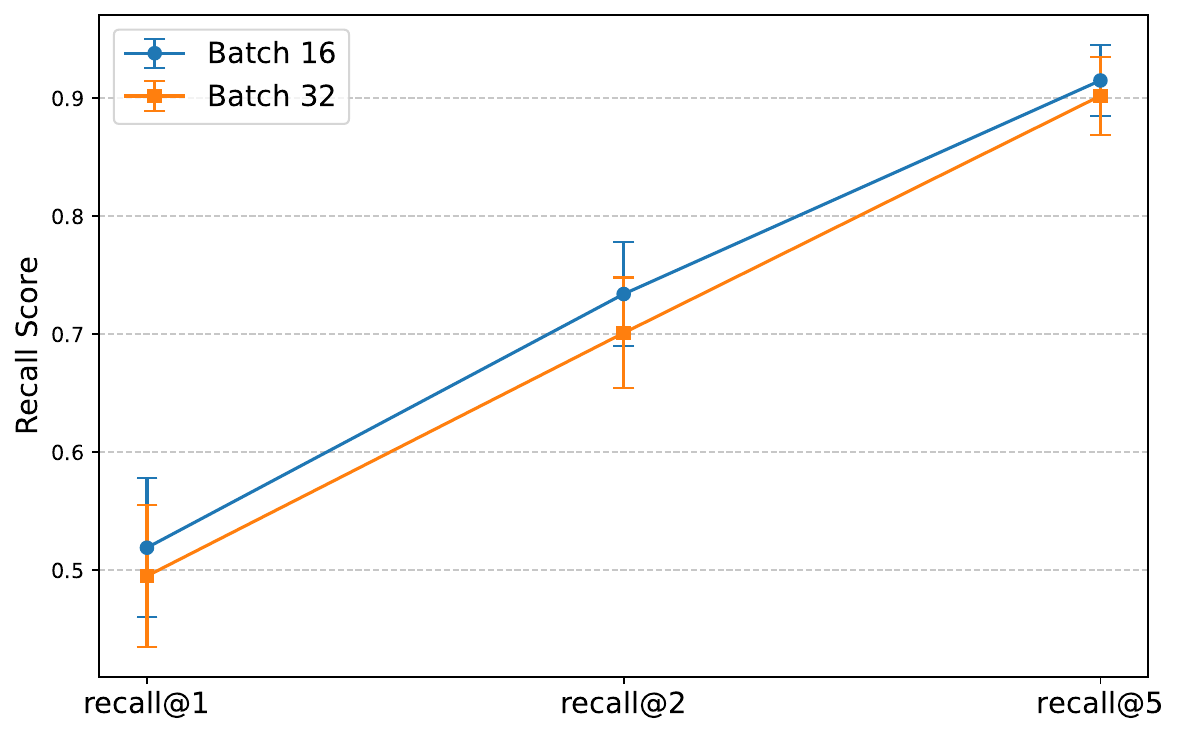}
    \caption{Effect of batch size on validation scores.}
    \label{fig:precision}
\end{figure}

We performed hyperparameter tuning for the number of epochs and batch size. After each training epoch, the model was evaluated on the validation set and the best-performing model was saved. For global contrastive learning, we experimented with two batch sizes (16 and 32). Figure \ref{fig:precision} illustrates the validation recall scores for these batch sizes. Since the validation scores for batch size 16 were higher than those for batch size 32, we selected 16 as the optimal batch size.

\subsection{Metrics}\label{sec:metrics}
These are the metrics we used for evaluation in current work:
\begin{enumerate}
    \item \textbf{precision@$k$} measures how many of the top $k$ items in a ranked list are relevant to a query.
    \begin{equation} \label{eq:5}
    \text{precision@k} = \frac{\text{\# of relevant items in top $k$}}{k},
    \end{equation}
    \item \textbf{recall@$k$} measures the proportion of relevant items that are successfully retrieved in the top $k$ items.
    \begin{equation}
    \text{recall@k} = \frac{\text{\# of relevant items in top $k$}}{\text{total number of relevant items}}. 
    \end{equation}
    \item \textbf{nDCG} (normalized discounted cumulative gain) is a ranking evaluation metric that measures the quality of a ranked list by considering both the relevance of items and their positions in the list.
    \item \textbf{MAP} (mean average precision) is the mean of the average precision (AP) scores in all queries. AP averages the precision scores at all ranks where a relevant item is retrieved for a single query.
\end{enumerate}

\subsection{Additional Case Study}
We conducted another case study showing the superiority of \method over the best baseline Trial2Vec. Similar to case study 1 (Table \ref{tab:case1}), we can see in case study 2 ( Table \ref{tab:case2}) that the top-1 trial retrieved by \method is more closely aligned with the query trial compared to the trial retrieved by Trial2Vec. Specifically, some important eligibility criteria, such as cancer stage and weight loss, match between the query trial and the retrieved trial by \method.

\begin{table*}[]
\resizebox{\textwidth}{!}{%
\begin{tabular}{@{}llll@{}}
\toprule
                                    & Query Trial                                                                                                                                                                                                                                                           & Trial2Vec                                                                                                                                                                                     & \method                                                                                                                                             \\ \midrule
NCTID                               & NCT01494558                                                                                                                                                                                                                                                           & NCT03800134                                                                                                                                                                                      & NCT00533949                                                                                                                                           \\ \hline
Title                               & \begin{tabular}[c]{@{}l@{}}Study of Etoposide, Cisplatin, \\ and Radiotherapy Versus Paclitaxel, \\ Carboplatin and Radiotherapy ...\end{tabular}                                                                                                                     & \begin{tabular}[c]{@{}l@{}}A Study of Neoadjuvant/Adjuvant \\ Durvalumab for the Treatment of \\ Patients With Resectable ...\end{tabular}                                                       & \begin{tabular}[c]{@{}l@{}}High-Dose or Standard-Dose \\ Radiation Therapy and Chemo-\\ therapy With or Without Cetuximab ...\end{tabular}            \\ \hline
Intervention                        & \begin{tabular}[c]{@{}l@{}}Chemoradiotherapy Regimen between PC \\ (paclitaxel 45mg/m2 weekly over 1hour \\ and carbplatin AUC =2mg/mL/min over \\ 30min weekly) and PE (etoposide \\ 50mg/m2 d1-5, 29-33 and cisplatin \\ 50mg/m2 d1,8,29 and 36 29-33)\end{tabular} & \begin{tabular}[c]{@{}l@{}}Drug: Durvalumab, Other: Placebo, \\ Drug: Carboplatin, Drug: Cisplatin, \\ Drug: Pemetrexed, Drug: Paclitaxel,\\  Drug: Gemcitabine, Procedure: Surgery\end{tabular} & \begin{tabular}[c]{@{}l@{}}Biological: Cetuximab, Drug: \\ Carboplatin, Drug: Paclitaxel, \\ Radiation: 60 Gy RT,\\  Radiation: 74 Gy RT\end{tabular} \\ \hline
Disease                             & Non-Small Cell Lung Cancer                                                                                                                                                                                                                                            & Non-Small Cell Lung Cancer                                                                                                                                                                       & Non-Small Cell Lung Cancer                                                                                                                            \\ \hline
\multirow{2}{*}{Important Criteria} & stage IIIA/IIIB NSCLC                                                                                                                                                                                                                                                 & Stage IIA to Stage IIIB                                                                                                                                                                          & stage IIIA/IIIB NSCLC                                                                                                                                 \\ \cline{2-4} 
                                    & lose weight \textless{}10\%                                                                                                                                                                                                                                           & -                                                                                                                                                                                                & lose weight \textless{}10\%                                                                                                                           \\ \hline
\end{tabular}
}
\caption{Case study comparing the retrieval performance of the SECRET and Trial2Vec.}
\label{tab:case2}
\end{table*}

\subsection{Example Trial Represented as a set of Q/A pairs}
Table \ref{tab:chickpea_pulao_study} shows an example trial (NCT06095622) after we represent it using Q/A pairs.  The current investigation limited evaluations of trial protocols to the following sections: \textit{title, disease, intervention, keywords, outcome} and \textit {eligibility criteria}. By \textit{outcome} in the paper, we mean only \textit{primary outcome measures}.

\begin{table*}[h!]
\centering
\begin{tabular}{|p{0.4\textwidth}|p{0.55\textwidth}|}
\hline
\textbf{Question} & \textbf{Answer} \\ \hline
What is the age requirement for participants? & 18 years \\ \hline
What type of diabetes is required for participation? & Type-2 \\ \hline
What is the dietary requirement for participants? & Chickpea rice pulao \\ \hline
What type of diet is not allowed for participants? & Vegan or keto \\ \hline
What are the drugs used? & Fenugreek Seeds and Indian Rennet \\ \hline
What is the disease treated in this trial? & Glucose Metabolism Disorders (Including Diabetes Mellitus) \\ \hline
What are the keywords? & Chickpea pulao \\ \hline
What is the title of the trial? & Formulation and Assessment of Chickpea Pulao Using Fenugreek Seeds and Indian Rennet for Improving Blood Glycemic Levels \\ \hline
What are the outcome measurements? & Improvement in blood glucose levels, Increase or decrease in postprandial glucose levels in mg/dL, 21 days \\ \hline
\end{tabular}
\caption{Trial NCT06095622 represented as a set of Q/A pairs.}
\label{tab:chickpea_pulao_study}
\vspace{-1em}
\end{table*}

\subsection{Additional Experiments on Partial Retrieval} 
\label{sec:partial_add}

Combining the title with additional sections (\textit{disease, intervention, keywords, outcome, eligibility criteria}) consistently improves both precision@1 and recall@1 compared to using the title alone. Among these combinations, integrating the title with the intervention section yields the highest precision@1 and recall@1.

\begin{figure}[!h]
    \centering
    \includegraphics[width=0.3\textwidth]{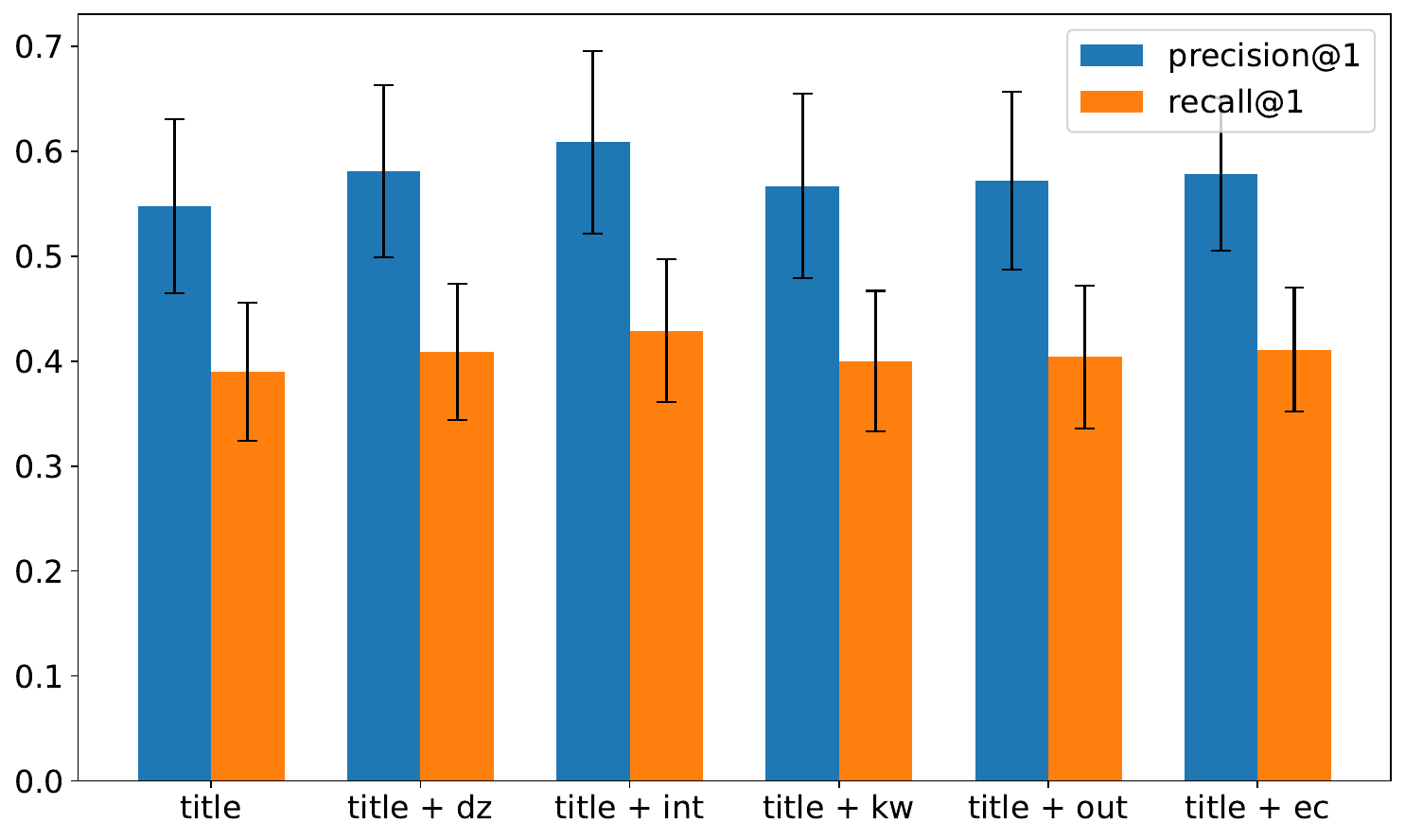}
    \caption{Performance of \method on the partial retrieval scenarios. We use different sections with title of the trial as queries to retrieve similar trials, including keyword kw, intervention int, disease dz, outcome out, eligibility criteria ec.}
    \label{fig:partial}
    \vspace{-1em}
\end{figure}

\subsection{Effect of Representing Trials with Q/A Pairs}\label{sec:ablation_plus}
\begin{figure}
    \centering
    \includegraphics[width=0.3\textwidth]{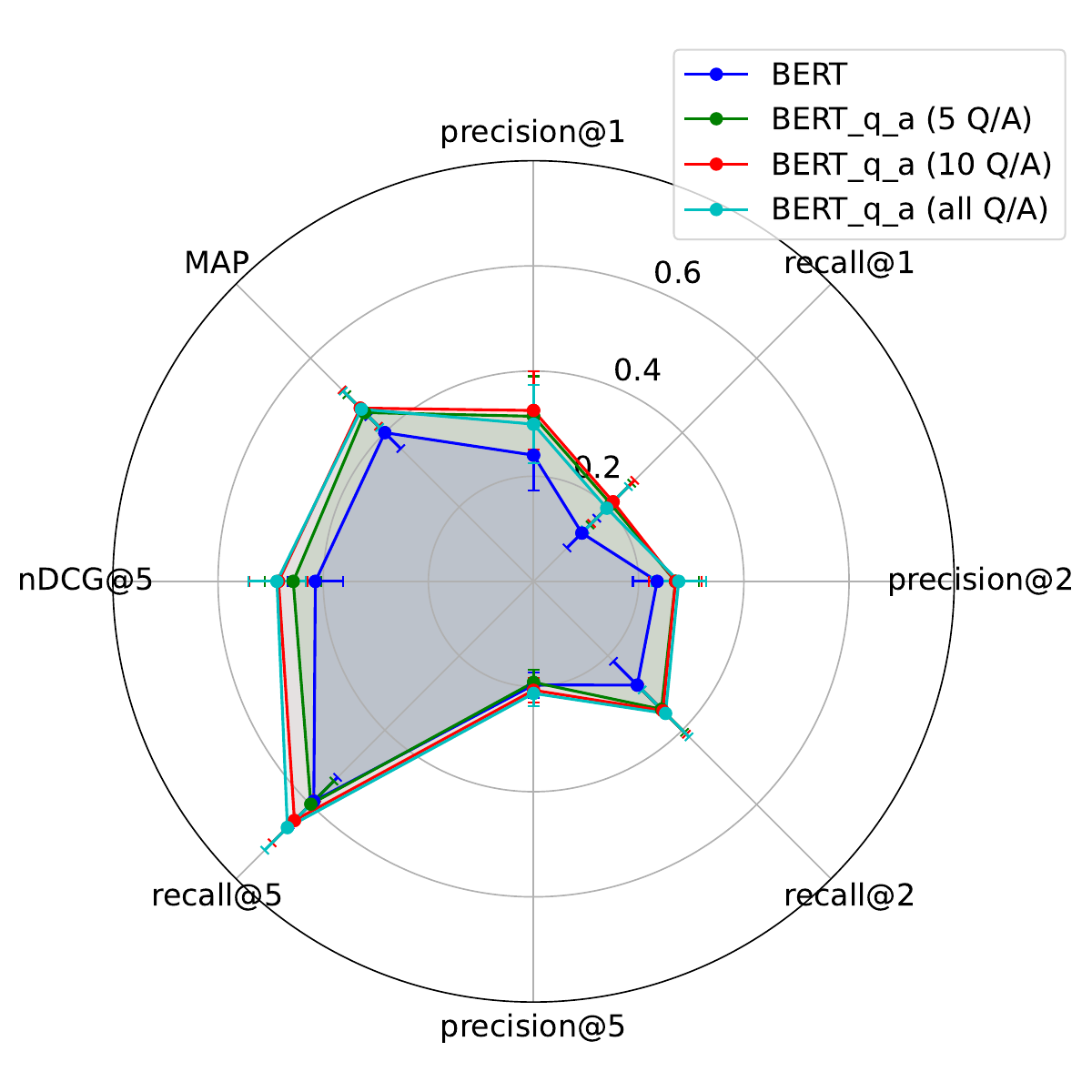}
    \caption{Performance evaluation of BERT and BERT\_q\_a models at different question-answer pair thresholds (5, 10, and all pairs) across various evaluation metrics: precision@1, recall@1, precision@2, recall@2, precision@5, recall@5, nDCG@5, and MAP.}
    \label{fig:bert_comparision}
\end{figure}

\begin{figure}[!h]
    \centering    \includegraphics[width=0.3\textwidth]{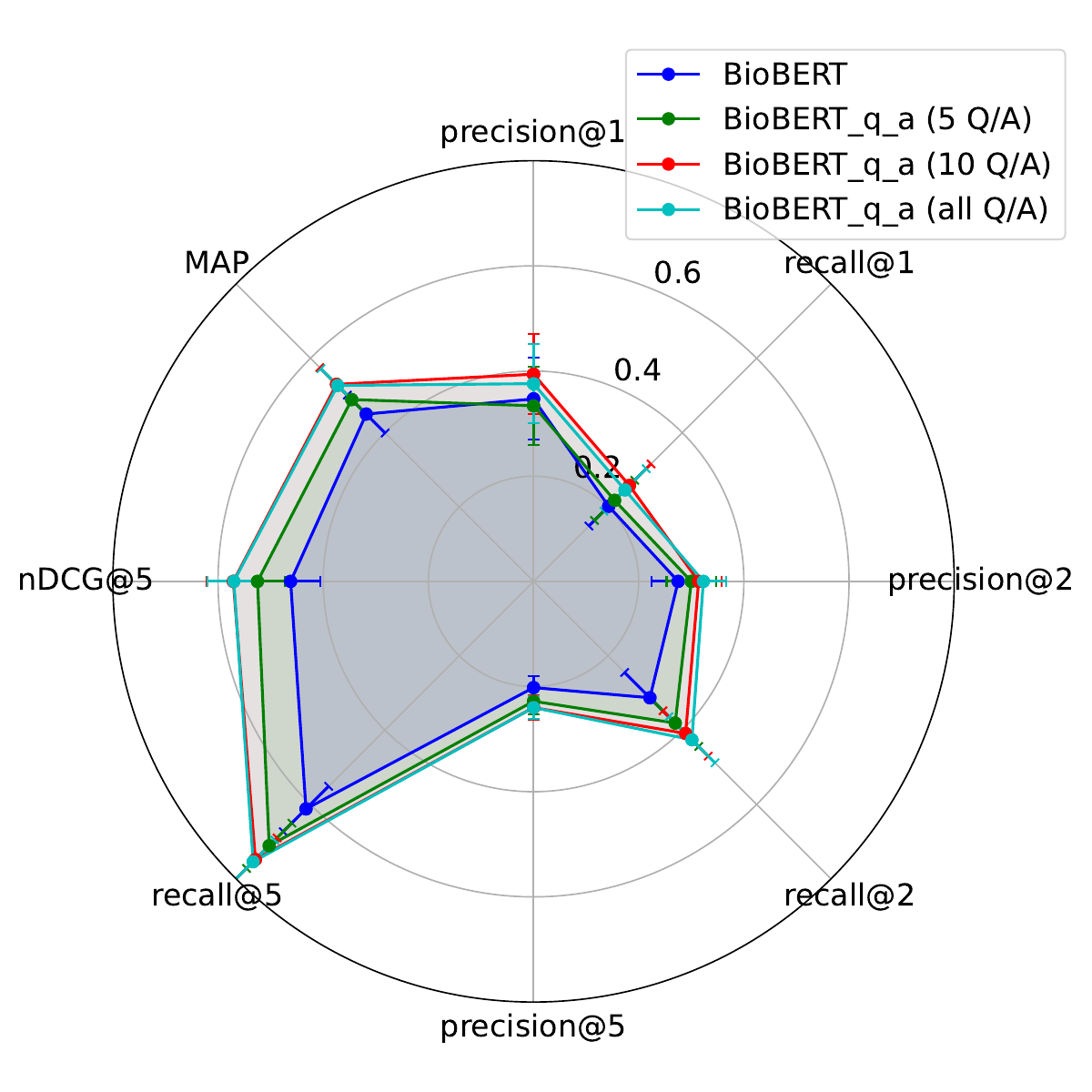}
    \caption{Performance evaluation of backbone BioBERT and BioBERT\_q\_a models at different question-answer pair thresholds (5, 10, and all pairs) across various evaluation metrics: precision@1, recall@1, precision@2, recall@2, precision@5, recall@5, nDCG@5, and MAP.}    \label{fig:biobert_comparision}
    \vspace{-1.5em}
\end{figure}

We performed some experiments to show the utility of using Q/A pairs as a way of representing the trial protocols. If we use whole sections of the trial protocols, some parts might get truncated depending on the length of the trial protocol and the method used to get embeddings. We show for BERT that representing trial protocols using Q/A pairs lead to significant improvement on retrieval scores. In Figure \ref{fig:bert_comparision}, BERT\_q\_a means that the trials are represented using Q/A pairs instead of the whole protocol. Similarly, in Figure \ref{fig:biobert_comparision}, BioBERT\_q\_a means that the trials are represented using Q/A pairs instead of the whole protocol. We experimented with 5, 10 and all Q/A generated by LLMs. 

\begin{table*}[!htb]
\resizebox{\textwidth}{!}{%
\begin{tabular}{@{}lllllllll@{}}
\toprule
                       & \textbf{precision@1} & \textbf{recall@1} & \textbf{precision@2} & \textbf{recall@2} & \textbf{precision@5} & \textbf{recall@5} & \textbf{nDCG@5} & \textbf{MAP} \\\midrule
\textit{Only local} & 0.495 $\pm$ 0.083 & 0.343 $\pm$ 0.065 & 0.386 $\pm$ 0.051 & 0.495 $\pm$ 0.064 & 0.249 $\pm$ 0.025 & 0.767 $\pm$ 0.057 & 0.627 $\pm$ 0.057 & 0.603 $\pm$ 0.055 \\
\textit{Only trial (full text)} & 0.560 $\pm$ 0.081 & 0.394 $\pm$ 0.064 & 0.463 $\pm$ 0.053 & 0.605 $\pm$ 0.068 & 0.275 $\pm$ 0.023 & 0.852 $\pm$ 0.050 & 0.716 $\pm$ 0.053 & 0.682 $\pm$ 0.054 \\
\textit{Only global (Q/A set)} & 0.589 $\pm$ 0.087 & 0.419 $\pm$ 0.069 & 0.481 $\pm$ 0.053 & 0.641 $\pm$ 0.063 & 0.278 $\pm$ 0.027 & 0.871 $\pm$ 0.051 & 0.739 $\pm$ 0.055 & 0.703 $\pm$ 0.054 \\
\textit{Predefined Q/A only} & 0.558 $\pm$ 0.084 & 0.401 $\pm$ 0.070 & 0.488 $\pm$ 0.056 & 0.644 $\pm$ 0.063 & 0.280 $\pm$ 0.027 & 0.877 $\pm$ 0.049 & 0.735 $\pm$ 0.051 & 0.701 $\pm$ 0.053 \\
\textit{Answers only} & 0.607 $\pm$ 0.080 & 0.438 $\pm$ 0.067 & 0.456 $\pm$ 0.055 & 0.604 $\pm$ 0.070 & 0.287 $\pm$ 0.023 & 0.890 $\pm$ 0.043 & 0.751 $\pm$ 0.048 & 0.710 $\pm$ 0.051 \\
\method & \textbf{0.647 $\pm$ 0.077} & \textbf{0.467 $\pm$ 0.063} & \textbf{0.508 $\pm$ 0.046} & \textbf{0.682 $\pm$ 0.061} & \textbf{0.297 $\pm$ 0.023} & \textbf{0.924 $\pm$ 0.034} & \textbf{0.796 $\pm$ 0.042} & \textbf{0.754 $\pm$ 0.044} \\ 
 \bottomrule
\end{tabular}}
\caption{Results of ablation study. The table presents precision, recall, nDCG, and MAP metrics, reported as mean $\pm$ standard deviation, with the best values highlighted in \textbf{bold}.}
\label{tab:ablation}
\end{table*}
\end{document}